\def\year{2019}\relax
\documentclass[letterpaper]{article} 
\usepackage{aaai19}  
\usepackage{times}  
\usepackage{helvet}  
\usepackage{courier}  
\usepackage{url}  
\usepackage{graphicx}  
\usepackage{amsfonts}
\usepackage{amsmath}
\usepackage{amssymb}

\begin{document}

\title{Exploiting Sentence Embedding for Medical Question Answering}


\author{Yu Hao$^{1*}$ , Xien Liu$^{1} \thanks{These two authors contribute equally to this work.}$, Ji Wu$^1$ \& Ping Lv$^2$\\
$^1$Department of Electronic Engineering, Tsinghua University, Beijing, China \\
haoy15@mails.tsinghua.edu.cn, \{xeliu, wuji\_ee \}@mail.tsinghua.edu.cn  \\
$^2$ Tsinghua-iFlytek Joint Laboratory, iFlytek Research, Beijing, China \\
luping\_ts@mail.tsinghua.edu.cn  \\
}
\maketitle

\begin{abstract}
Despite the great success of word embedding, sentence embedding remains a not-well-solved problem. In this paper, we present a supervised learning framework to exploit sentence embedding for the medical question answering task. The learning framework consists of two main parts: 1) a sentence embedding producing module, and 2) a scoring module.  The former is developed with contextual self-attention and multi-scale techniques to encode a sentence into an embedding tensor. This module is shortly called  Contextual self-Attention Multi-scale Sentence Embedding (CAMSE).  The latter employs two scoring strategies: Semantic Matching Scoring (SMS) and Semantic Association Scoring (SAS).  SMS measures similarity while SAS captures association between sentence pairs: a medical question concatenated with a candidate choice, and a piece of  corresponding supportive evidence.  The proposed framework is examined by two Medical Question Answering(MedicalQA) datasets which are collected from real-world applications: medical exam and clinical diagnosis based on electronic medical records (EMR).   The comparison results show that our proposed framework achieved significant improvements compared to competitive baseline approaches.  Additionally, a series of controlled experiments are also conducted to illustrate that the multi-scale strategy and the contextual self-attention layer play important roles for producing effective sentence embedding, and the two kinds of scoring strategies are highly complementary to each other for question answering problems.
\end{abstract}

\section{Introduction}
  Embedding  learning in word-level has achieved much progress\cite{bengio2003neural,mikolov2013distributed,mikolov2013efficient,pennington2014glove}  and the pre-trained word embeddings have been almost a standard input to a certain deep learning framework for solving upstream applications, such as reading comprehension tasks \cite{raison2018weaver,wang2018multi,zhang2018medical,chen2017reading,Cheng2016Long,dhingra2016gated,seo2016bidirectional}.  However, learning embeddings at sentence/document level is still a very difficult task, not well solved at present. 
The study of sentence embedding runs along the two lines: 1) exploiting semantic/linguistic properties  \cite{zhu2018exploring,baroni2018you}  obtained within sentence embeddings, and 2) designing  learning methods to produce effective sentence embeddings.  All of the learning methods can be generally categorized into two groups: 1) obtaining universal sentence embeddings with an unsupervised learning framework \cite{hill2016learning,kiros2015skip,le2014distributed}, and 2) producing  task-dependent 
sentence embeddings with a supervised learning framework \cite{palangi2016deep,tan2016improved,feng2015applying,Cheng2016Long,Lin2017A}.

Though plenty of successful deep learning models are built at word level (word embedding),  there are still some different and valuable merits obtained within sentence embedding. 
For example, most reading comprehension models calculate a pairwise similarity at word level to extract keywords in the answer. However, these fine-grained models maybe misled under certain circumstances, such as long paragraph with lots of single noisy words which are similar to those words that appear in the question but unrelated to the question answering.
Furthermore, models built on sentence embeddings sometimes can be more interpretable. 
For example, we can encode a sentence into several embeddings to capture different semantic aspects of the sentence. As we known,  sometimes interpretability becomes more crucial for  certain real applications, such as tasks from the medical domain.

In this paper, we focus on developing supervised sentence embedding learning framework for solving medical question answering problems. To maintain model interpretation, we also adopt the self-attention structure proposed by \cite{Lin2017A} to produce sentence embeddings.  The only difference is that a contextual layer is used in conjunction with the self-attention.  Under certain circumstances, the valuable information resides in a unit whose size is between word and sentence. Take medical text for an instance, a large amount of the medical terminologies are entities consist of several sequential words like \emph{Acute Upper Respiratory Infection}. It requires a flexible scale between word and sentence-level to encode such sequential words as a single unit and assign words in the unit with similar attention, instead of treating them like a bag of unrelated words, which can be misled easily by noisy words in long paragraphs when computing pairwise word similarities. For example, sentences that include \emph{Acute Gastroenteritis} and \emph{Acute Cholecystitis} may be considered to some extent related to question that describes \emph{Acute Upper Respiratory Infection} because \emph{Acute} appears in all of them, even though these sentences concentrate on totally different diseases. Therefore we propose contextual self-attention and multi-scale strategy to produce sentence embedding tensor that captures multi-scale information from the sentence. The contextual attention detects meaningful word blocks(entities) and assigns words in the same block with similar attention value. The multi-scale allows the model to directly encode sequential words as an integral unit and extracts informative entities or phrases. The contextual attention is a soft assignment of attention value, while the multi-scale strategy is a hard binding of sequential words.  Even though being able to preserve more information by producing a tensor, \citeauthor{Lin2017A} simply calculate similarities between corresponding semantic subspaces and fail to capture the association between different subspaces. In an attempt to fully exploit the abundant information lies in the tensor, we propose two scoring strategies: Semantic Matching Scoring(SMS) and Semantic Association Scoring(SAS).

In the rest of this paper, we will first define the medical question answering task and introduce two datasets. Then, the supervised sentence embedding learning framework (consisting of sentence embedding producing module CAMSE and scoring module) are introduced, and a series of comparison results and some crucial analysis are presented.


\section{MedicalQA Task Description}

Here, we define the MedicalQA task with three components:

\begin{itemize}
\item[-] Question: a short paragraph/document in text describing a medical problem.
\item[-] Candidate choices $ $: multiple candidate choices are given
for each question, and only one is the correct answer.
\item[-] Evidence documents: for each candidate choice, a collection of short documents/paragraphs\footnote{In the rest of this paper, we will not specifically differentiate sentences from documents/paragraphs. These terms can be  used interchangeably. } is given as evidence to support the choice as the right answer.
\end{itemize} 
The goal of MedicalQA is to determine the correct answer based on corresponding evidence documents with an appropriate scoring manner. 
\begin{equation}
\label{equ_qac}
	(Q, \{c_1, c_2,... c_{n_c} \}, \{ D_1, D_2, ..., D_{n_c} \} )\to c\ast,
\end{equation}
where $n_c$ is the number of candidate choices for each question, $c_i$ is the  $i$th candidate choice, and $D_i=\{d_{i1}, d_{i2}, ...d_{in_e}\}$ is the set of evidence documents for the choice $c_i$, where $i=1, 2,...,n_c$ is the index of candidate choice, and $n_e$ is the number of evidence documents for each choice. 

   In the rest of this section, we will introduce two kinds of medical question answering problems, which are from real-world applications and can be transformed into MedicalQA task as defined in formula (\ref{equ_qac}). The first task comes from a medical exam: the General Written Test (GWT) of National Medical Licensing Examination in China ({{\bf NMLEC}), and the second task is Clinical Diagnosis based on Electronic Medical Records({\bf CD-EMR}). 

\subsection{MedicalQA\#1:NMLEC}

\subsubsection{Data source}
NMLEC is an annual certification exam which comprehensively evaluates doctors'  medical knowledge and clinical skills. The General Written Test part of NMLEC consists of 600 multiple choice questions. Each question is given 5 candidate choices (one example is presented in Fig.  \ref{fig_example_nmlec}), and only one of them is the correct answer. The exam examines more than 20 medical subjects 

\begin{figure}[htb!]    
\centering    
    \center{\includegraphics[scale=0.35] {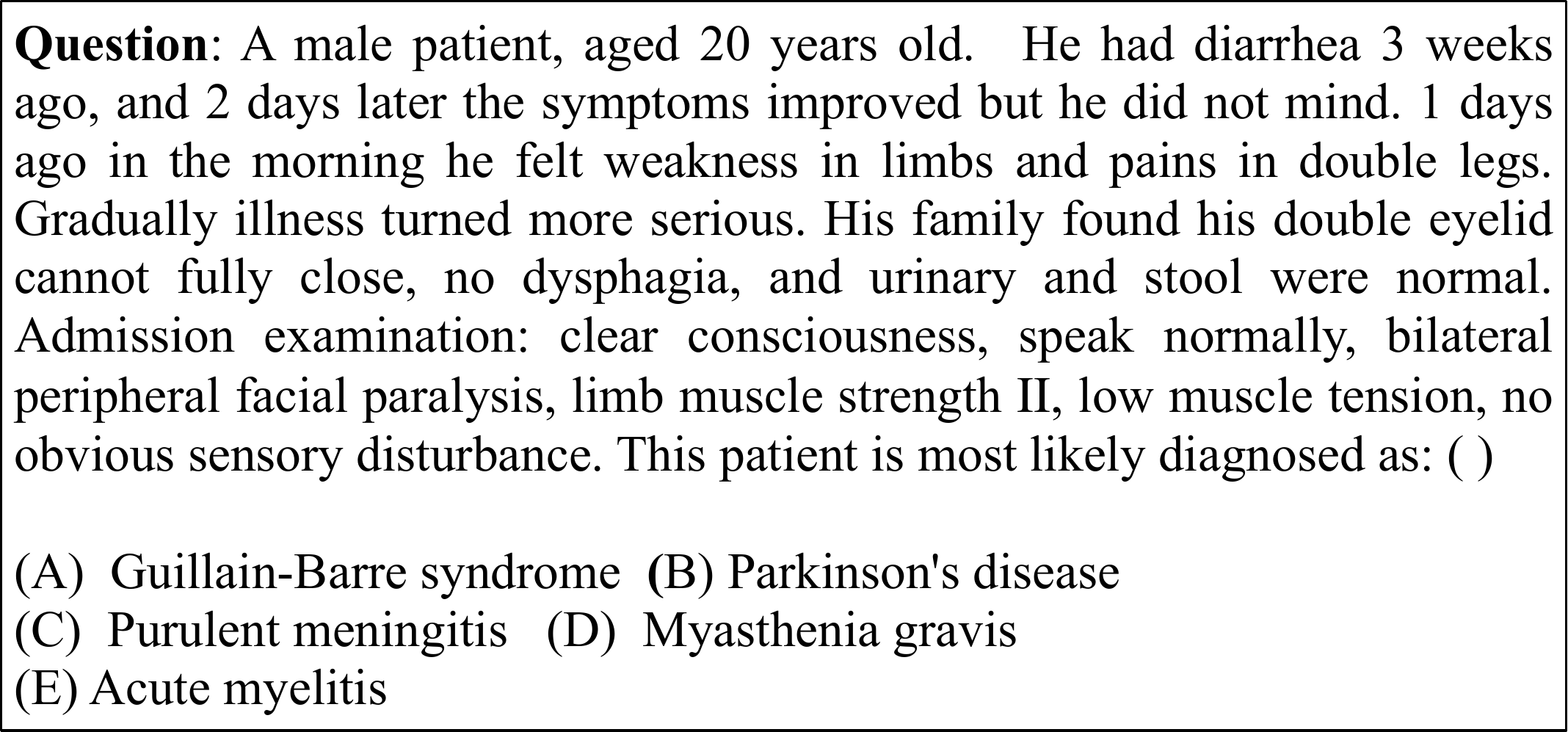}}
    \caption{ An example question from the General Written Test part of NMLEC.}  
    \label{fig_example_nmlec}    
\end{figure}


\subsubsection{Training/test set}
We collected 10 suites of the exam, totally 6,000 questions, as the test set.
To avoid test questions appearing in the training set with minor variance, we dropped the training questions which are very similar to the questions from the test set, resulting in totally 250,000 medical questions as the training set.  The similarity of two questions is measured by comparing Levenshtein distance\cite{Levenshtein1966Binary} with a threshold ratio of 0.8.

\subsubsection{Evidence documents}
The General Written Test of NMLEC mainly examines medical knowledge from medical textbooks. 
Therefore, we first collected totally more than 30 publications (including textbooks, guidebooks etc.) as evidence source.  Then we produced evidence documents from the evidence source with a text retrieval system built upon Apache Lucene with BM25 ranking.

\subsection{MedicalQA\#2:CD-EMR}

\subsubsection{Data source}

A large amount of electronic medical records(EMRs) are collected from the outpatient department of basic level hospitals. An example of the EMRs sample is listed in Fig.\ref{fig_example_clinical}. The EMRs mainly consists of three parts:
\begin{itemize}
	\item [-] Chief complaint: a very brief description of the patient's illness symptom.
	\item [-] History of present illness\footnote{History of Present Illness, commonly abbreviated HPI, is also termed History of Presenting Complaint (HPC) in medical domain.}: a more detailed interview (comprehensive description to the patient's illness) prompted by the chief complaint or presenting symptom. 
	\item [-] Disease code: a code consisting of a combination of uppercase and numbers with a length of four. Each code indicates only one disease. 
\end{itemize}

\begin{figure}[htb!]        
    \center{\includegraphics[width=5cm] {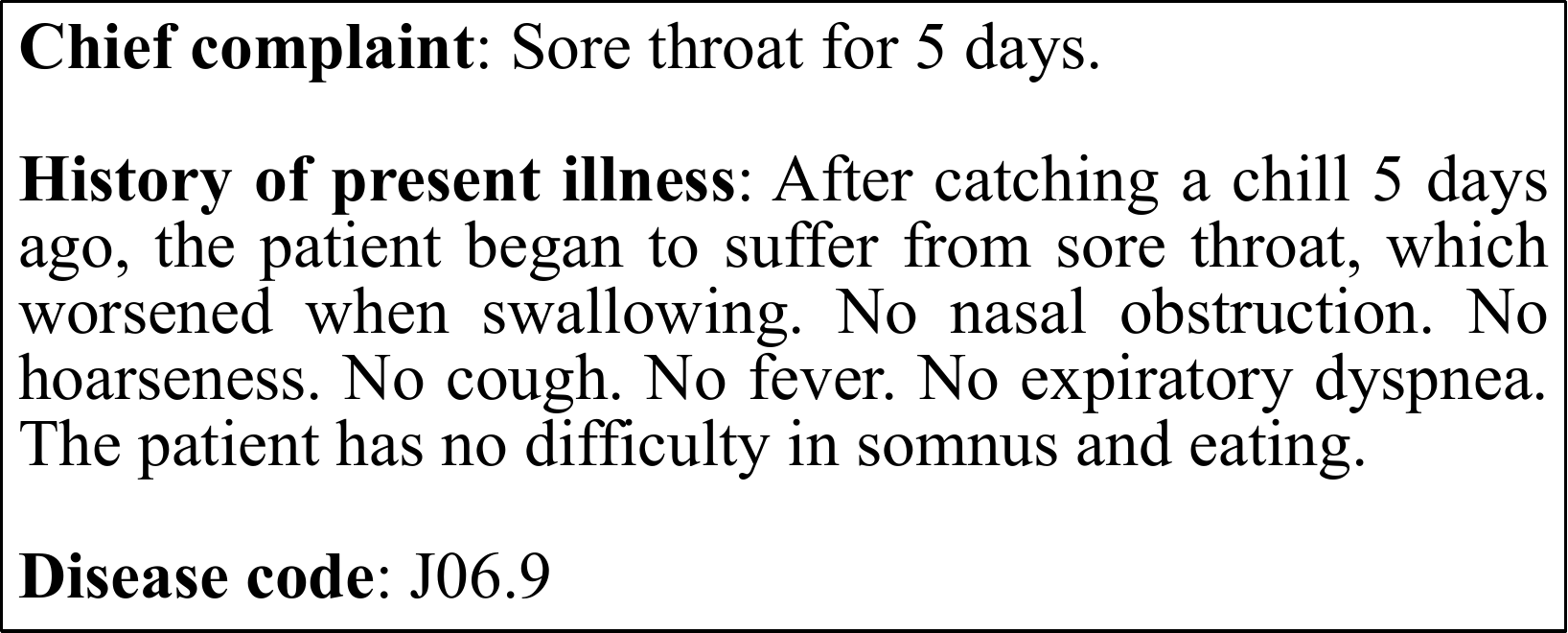}}
    \caption{ An example EMR data.}  
    \label{fig_example_clinical}    
\end{figure}

\subsubsection{Questions \& candidate choices}
 To transform the EMRs data (see Fig.\ref{fig_example_clinical}) into a standard MedicalQA task defined above, here we concatenate each chief complaint and its corresponding history of present illness, and treat the concatenation as a medical question. There are totally 98 disease codes in our data.
 
\subsubsection{Training/test set}

We collected EMRs data during a long period from the outpatient department of basic level hospitals
as the training set, and the next period data collected from the same hospitals are used as test set. 
Here, we use the next period data as the test set.
The main purpose is to make the problem more suitable for real applications (learning from historical data, but predicting over present/future data).  All training/test EMRs data are transformed into MedicalQA questions via the method mentioned above. The training set has 75265 items, and the test set has 16551 items.



\subsubsection{Evidence documents}
Since MedicalQA\#2 are collected from real-world EMRs data, the question description varies significantly due to the diversity of human doctors' writing styles and deviates from textbook styles as well.  Using a text retrieval system to retrieve evidence documents as done in MedicalQA\#1 is not a good choice (We have a try, but the results are terrible).  Here, we selected similar question samples from training set as evidence documents.  All training questions are used to train a simple LSTM-MLP classifier. The output of LSTM-MLP is treated as a representation of the problem and we use it to select nearest neighbors of each problem as their supportive documents for each disease.


\section{The Framework}

According to the  MedicalQA task defined above (\ref{equ_qac}),  the key to determining the correct answer from many candidate choices is to evaluate the supportive degree of an evidence document to the corresponding candidate choice.  In this study, we consider the supportive degree into two aspects: how semantic similarity between the sentence pairs (question concatenated with a candidate choice, and the corresponding evidence document), and how association across the sentence pairs.  The former is degreed with Semantic Matching Scoring ({\bf SMS})  and the latter is measured with  Semantic Association Scoring ({\bf SAS}).  In the rest of this section, we will first introduce the sentence embedding producing module CAMSE and then present the scoring module consisting of SMS and SAS. 

\begin{figure*}[htb!]        
    \center{\includegraphics[scale=0.35]  {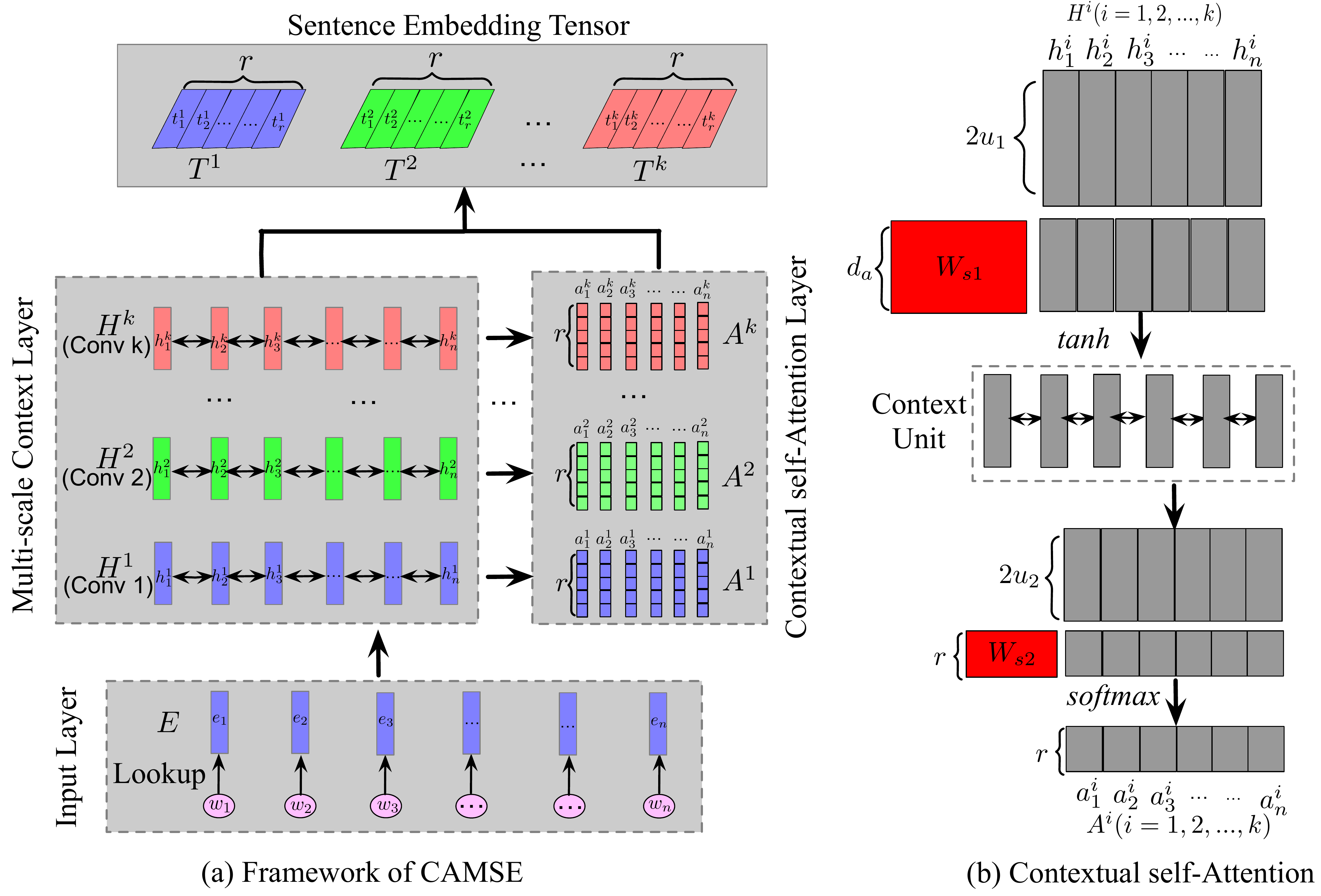}}
    \caption{The framework of sentence embedding learning: CAMSE}  
    \label{fig_CAMSE}    
\end{figure*}

\subsection{CAMSE}
The sentence embedding producing module  CAMSE  is presented in Fig.\ref{fig_CAMSE}.  We will introduce its details layer by layer. 

\subsubsection{Input layer \& multi-scale context layer}
For each word of the input sentence, we lookup its embedding from pre-trained embedding matrix. Then we implement a multi-scale convolution operation on word embeddings with variable window size $i$($i=1,2,3,...,k$). (see Fig.\ref{fig_CAMSE} (a)).The variation of granularity enables the model to not only process single words, but also bind sequential $i$ words as an integral representation for the potential terminology of entities or phrases, like $\emph{chronic\ bronchitis}$. The outputs of convolution with different window sizes are processed separately with different bidirectional LSTM networks to generate semantic encodings.

\subsubsection{Contextual self-attention layer}
Similar to previous self-structured sentence embedding model\cite{Lin2017A}, at each scale a sentence $H^i=(h^i_1, h^i_2, h^i_3,...,h^i_n)$ with variable length $n$ is encoded into a fixed-shaped 2-D feature matrix. The multi-attention mechanism attends to different semantic components of the sentence, and preserve more information than a single vector. We encode each word in the sentence into a r-dimensional attention vector, representing its significance in corresponding semantic sub-spaces. 

As Fig.\ref{fig_CAMSE} (b) shows, we first use an unbiased 1-layer feed-forward network to compress the word representation. $W^i_{s1}\in\mathbb{R}^{2u_1\times d_a}$, $d_a$ is the size of hidden state and $u_1$ is the one-direction output size of Bi-LSTM in the previous layer.

\begin{equation}
    M^{i,1}=\text{tanh}(H^iW^i_{s1})
\end{equation}

The hidden states $M^{i,1}=(m^{i,1}_1,m^{i,1}_2,\ldots,m^{i,1}_n)$  are processed with a 1-layer Bidirectional LSTM to integrate context information. 
Although the LSTM network in multi-scale context layer has contained dynamic information, we still adopt another Bi-LSTM layer here to separate the function of two Bi-LSTM networks. The first Bi-LSTM layer concentrates on semantic encoding which is further utilized when producing sentence embeddings. The output embeddings of sequential words in entities can vary significantly after the first Bi-LSTM layer, in order to preserve the diverse semantic information in a sentence. The second layer, on the other hand, focuses on detecting the meaningful word blocks and assigns words in them with similar attention values. The contextual information is incorporated so that the attention layer can better capture word blocks and treat the words in a block equally, even though their semantic embeddings might vary drastically.

\begin{equation}
    m_t^{i,2}=\text{Bi-LSTM}(m_{t-1}^{i,2}, m_{t}^{i,1})
\end{equation}

$M^{i,2}=(m^{i,2}_1,m^{i,2}_2,\ldots,m^{i,2}_n)$, where $M^{i,2}\in\mathbb{R}^{n\times 2u_2}$, 
$u_2$ is the one-direction output size of Bi-LSTM in the context unit. The outputs of Bi-LSTM network at steps are then encoded with a one-layer feed-forward network. The softmax is performed along the first dimension to ensure the attention of words in a semantic sub-space sums to 1.

\begin{equation}
    A^i=\text{softmax}(M^{i,2}W^i_{s2})
\end{equation}

$W^i_{s2}\in\mathbb{R}^{2u_2\times r}$, r is the number of semantic sub-spaces. $A^i\in\mathbb{R}^{n\times r}$ is the attention matrix.$A^i=(a^i_1, a^i_2, a^i_3,...,a^i_n)$, where $i=1,2,3,...,k$, each element $a^i_j\in R^{r\times 1} (j=1,2,3,..,n)$ is an attention vector.


\subsubsection{Sentence embedding tensor}
The attention matrix $A^i$ is used as summation weights to summarize representations in a sentence. (see the upper part of Fig.\ref{fig_CAMSE} (a))

\begin{equation}
    T^i=(A^i)^TH^i
\end{equation}

$T^i=(t^i_1, t^i_2, ..., t^i_r)$, where $t^i_j \in \mathbb{R}^{1\times 2u_1}$ is an embedding vector, and $r$ is the number of semantic sub-spaces. 

$T=(T^1, T^2, ..., T^k)$ is the sentence embedding tensor generated by CAMSE(Contextual self-attention Multi-scale Sentence Embedding). The tensor $T\in\mathbb{R}^{k\times r \times 2u_1}$ automatically aligns information from the sentence in k scales and r aspects.

\subsection{Scoring module}

\begin{figure*}[htb!]        
    \center{\includegraphics[scale=0.3]  {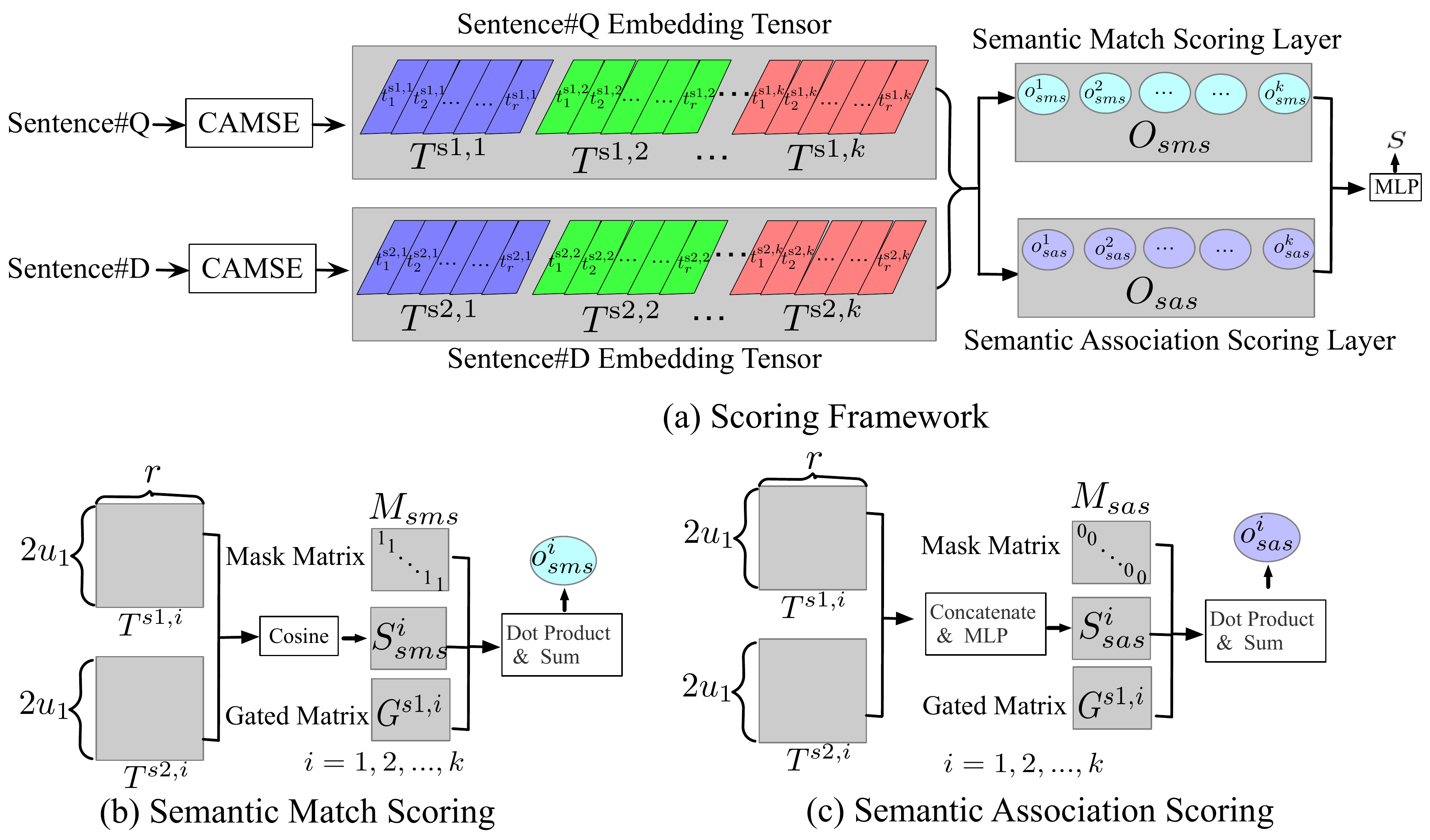}}
    \caption{The framework of scoring based on sentence embedding.}  
    \label{fig_score}    
\end{figure*}

Given a pair of sentences (Q, D), we first apply CAMSE to attain sentence embedding tensors $T^{s1}, T^{s2}$ respectively for question and document.(see Fig.\ref{fig_score} (a)). Then a scoring function takes the tensor pair ($T^{s1}, T^{s2}$) as input, computing a scalar score S as supporting degree of the document.

We propose two approaches of scoring, the Semantic Matching Scoring({\bf SMS}) and Semantic Association Scoring({\bf SAS}).(see Fig. \ref{fig_score} (a)). These two methods can be utilized together to boost the performance. A question-dependent gated matrix $G^{s1,i}\in\mathbb{R}^{r\times r}$ together with masks $M_{sms}\in\mathbb{R}^{r\times r}$ and $M_{sas}\in\mathbb{R}^{r\times r}$ control the information flow of two methods. As shown in Fig.\ref{fig_score} (a). The scores of two approaches from all k scales are aggregated with a 1-layer MLP to predict a scalar score S, where $\mathbf{w_s}\in\mathbb{R}^{2k}$.

\begin{equation}
    S=\mathbf{w_s^T}[O_{sms}^1, \ldots, O_{sms}^k, O_{sas}^1, \ldots, O_{sas}^k]
\end{equation}

Here, the scalars $O_{sms}^i$ and $O_{sas}^i$ are the outputs of Semantic Matching Scoring and Semantic Association Scoring of the $i$th scale. 

\subsubsection{SMS: Semantic Matching Scoring}
The $u$th column of two embedding tensors $T^{s_1, i}_u$ and $T^{s_2, i}_u$, are aligned to the same semantic sub-spaces. We compute a cosine similarity for each pair of semantic sub-space embedding columns.(see Fig.\ref{fig_score} (b))

\begin{equation}
    S_{sms}^i(u,u)=\frac{T^{s_1, i}_u\cdot T^{s_2, i}_u}{\|T^{s_1, i}_u\|\|T^{s_2, i}_u\|}
\end{equation}


\subsubsection{SAS: Semantic Association Scoring}
As for different columns of two embedding tensors, we cannot simply compute their cosine similarities because different semantic sub-spaces are not aligned. However, we can utilize the inter-semantic relationship to exploit associations between different semantic sub-spaces. We concatenate two embeddings and send them into a 1-layer MLP to measure the correlation between these two semantics.(see Fig.\ref{fig_score} (c)). The MLP outputs a scalar value for each semantic pair $(u,v)$. Different semantic pairs hold different sets of parameters
$\mathbf{w_{uv}}\in\mathbb{R}^{4u_1}$. 

\begin{equation}
    S_{sas}^i(u,v)=\text{sigmoid}(\mathbf{w_{uv}^T}[T^{s_1, i}_u, T^{s_2, i}_v])
\end{equation}


The intuition is based on the fact that though words in different semantics are not aligned, they may frequently co-occur in the data. The fully-connected layer takes advantage of the co-occurrence, as a complement to the SMS approach. Take the clinical data for an example, descriptions for a disease contains several aspects. Some focus on particular symptoms while others narrate what triggers the disease. The cause of disease \emph{catch a cold} has an association with symptom \emph{cough}. The inter-semantics scoring successfully represents the association between different semantic sub-spaces(symptom and pathogeny in this case).

\subsection{Gated matrix}
We use a matrix gate to determine which semantic pairs play pivotal roles in answer predicting, and the semantic pairs containing irrelevant information should be discarded.


\begin{equation}
    T^{s1,i}_{flat}=\text{flatten}(T^{s1,i})
\end{equation}
\begin{equation}
    G_{flat}=\text{sigmoid}({W_{g2}}tanh(W_{g1}T^{s1,i}_{flat}))
\end{equation}
\begin{equation}
    G=\text{reshape}(G_{flat}, [r,r])
\end{equation}

The mask matrices $M_{sms}$ and $M_{sas}$ respectively get diagonal and non-diagonal part of the matrix.

\begin{equation}
    O_{sms}^i=\text{sum}(S_{sms}^i \odot M_{sms} \odot G^{s1,i})
\end{equation}
\begin{equation}
    O_{sas}^i=\text{sum}(S_{sas}^i \odot M_{sas} \odot G^{s1, i})
\end{equation}	


\section{Experiments}


\subsection{Evaluation protocol}
All the models should generate a score $S$ for a $(Statement, Document)$ pair. We sum scores of all documents belong to a candidate answer as its reliability and select the one with the highest reliability as the correct answer.
\begin{equation}
	\label{equ_sum_all_doc}
	S_i=\sum^{n_e}_{e=1}{S_{ie}}
\end{equation} 
\begin{equation}
	\label{equ_select_true}
	c\ast=\mathop{\arg\max}_{i} \ S_i \ (i=1,2,\ldots,n_c)
\end{equation}

For those sentence embedding models, we use them to generate a sentence embedding vector separately for question(statement) and the documents, which can be considered as a siamese network. We compute cosine similarity as the score for each pair. 
The machine comprehension models, such as R-Net, are intended for datasets like SQuAD\cite{2016arXiv160605250R} that requires an answer span in a paragraph. We modify these models by replacing the output layer with MLP layer that outputs a scalar score as the supportive degree of the document to the statement.

We report question answering accuracy on the test set. The answer predicted by model and the true answer are denoted as $c_{\ast}^{i}$ and $c_{true}^{i}$ for the $i$th question. The indicator function $\mathbb{I}(x)$ is 1 when x is True, and 0 otherwise. 
\begin{equation}
Accuracy = \frac{\sum_{i=1}^{N}\mathbb{I}({c_{\ast}^{i} = c_{true}^{i}})}{N}
\end{equation}

\subsection{Pre-trained word embeddings}
Word embeddings in the input layer are trained on their corresponding medical text corpus using skip-gram \cite{mikolov2013efficient}. 
In MedicalQA\#1, Word embeddings are trained on all collected medical textbooks and examination questions in train set; In MedicalQA\#2, Word embeddings are trained on all collected EMRs data. 
The embedding's dimension is set to 200 for MedicalQA\#1 and 100 for MedicalQA\#2. Unseen words during testing are mapped to a zero vector.

\subsection{Model settings}
To save training time on the GPU, we truncate all evidence documents and questions to no more than 100 words for MedicalQA\#1 and 70 words for MedicalQA\#2. 
For each candidate choice, only top 10 evidence documents are used to calculate the supportive score. The Bi-directional LSTM in the context layer has a dimension of 128.
 The size of attention encoding hidden state $d_a$ (see Fig.\ref{fig_CAMSE}(b)) is 100. The number of semantics, $r$, is 15. Without any specification, in the multi-scale context layer of  CAMSE framework, the size of convolution is 1,2,and 3. 

\subsection{Training}
We put a softmax layer on top of candidate scores and use cross-entropy as loss function. Our model is implemented with Tensorflow \cite{abadi2016tensorflow}. We use Adam optimizer with exponential decay of learning rate and a dropout rate of 0.2 to reduce overfit, and the batch size is 10.

\subsection{Results and analysis}

We conduct a comparison of our model CAMSE with competitive baseline approaches, including sentence embedding models such as LSTM+DSSM\cite{Palangi2015Deep}, LSTMN\cite{Cheng2016Long}, and Self-Attention \cite{Lin2017A}; and including some famous reading comprehension models, such as R-Net\cite{rnet}, Iterative Attention\cite{sordoni2016iterative}, Neural Reasoner\cite{peng2015towards}, and SeaReader\cite{zhang2018medical}. The comparison results over the two kinds of MedicalQA tasks are presented in Table \ref{tab-compare-total}.  From the results, we can see that our presented model achieve remarkable gains than other sentence embedding models (LSTM+DSSM, LSTMN,  Self-Attention) and is also superior to the competitive reading comprehension models (SeaReader, R-Net, Iterative Attention, and Neural Reasoner). 

The performances of sentence-level models (LSTM+DSSM, LSTMN, Self-Attention) are generally poorer than word-level machine comprehension models(SeaReader, R-Net, Iterative Attention, and Neural Reasoner), indicating the difficulty of solving question answering problems with sentence embeddings. Our sentence-level approach, however, achieves even better performance compared with machine comprehension models.


\begin{table}[h]
\normalsize
    \centering
    \caption{Experimental results comparison of our  CAMSE model with  other baseline approaches.}
    \begin{tabular}{ccc}
        \hline
         Models & MedicalQA\#1 & MedicalQA\#2 \\ 
        \hline
        LSTM-DSSM  & 44.1  & 81.5\\
        LSTMN & 45.0 & 81.6\\
         Self-Attention & 65.2 & 78.5\\
         CAMSE & {\bf 73.6}  & {\bf 84.3} \\ 
        \hline
        Neural Reasoner & 52.5 & 81.1 \\ 
        Iterative Attention & 58.7  & 82.1 \\ 
        R-Net  & 63.7  & 82.4 \\ 
        SeaReader & 71.8  & 82.4 \\ 
        \hline
    \end{tabular}
    \label{tab-compare-total}
\end{table}

\subsection{Contextual self-attention}
Fig.\ref{context_eg} shows an example of how the contextual self-attention works. The first attention attends to information related with $\emph{"neck mass"}$; the second attention promotes to represent question type $\emph{"the most meaningful inspect for diagnosis"}$; while the third mainly focuses on $\emph{"fine-needle aspiration cytology"}$, an inspect method of thyroid. We also noticed that the sequential words in terminologies or phrases are equally assigned with high attention value, indicating that they are encoded as a whole unit via the contextual self-attention mechanism.  

\begin{figure}[htb!]        
    \center{\includegraphics[scale=0.5]  {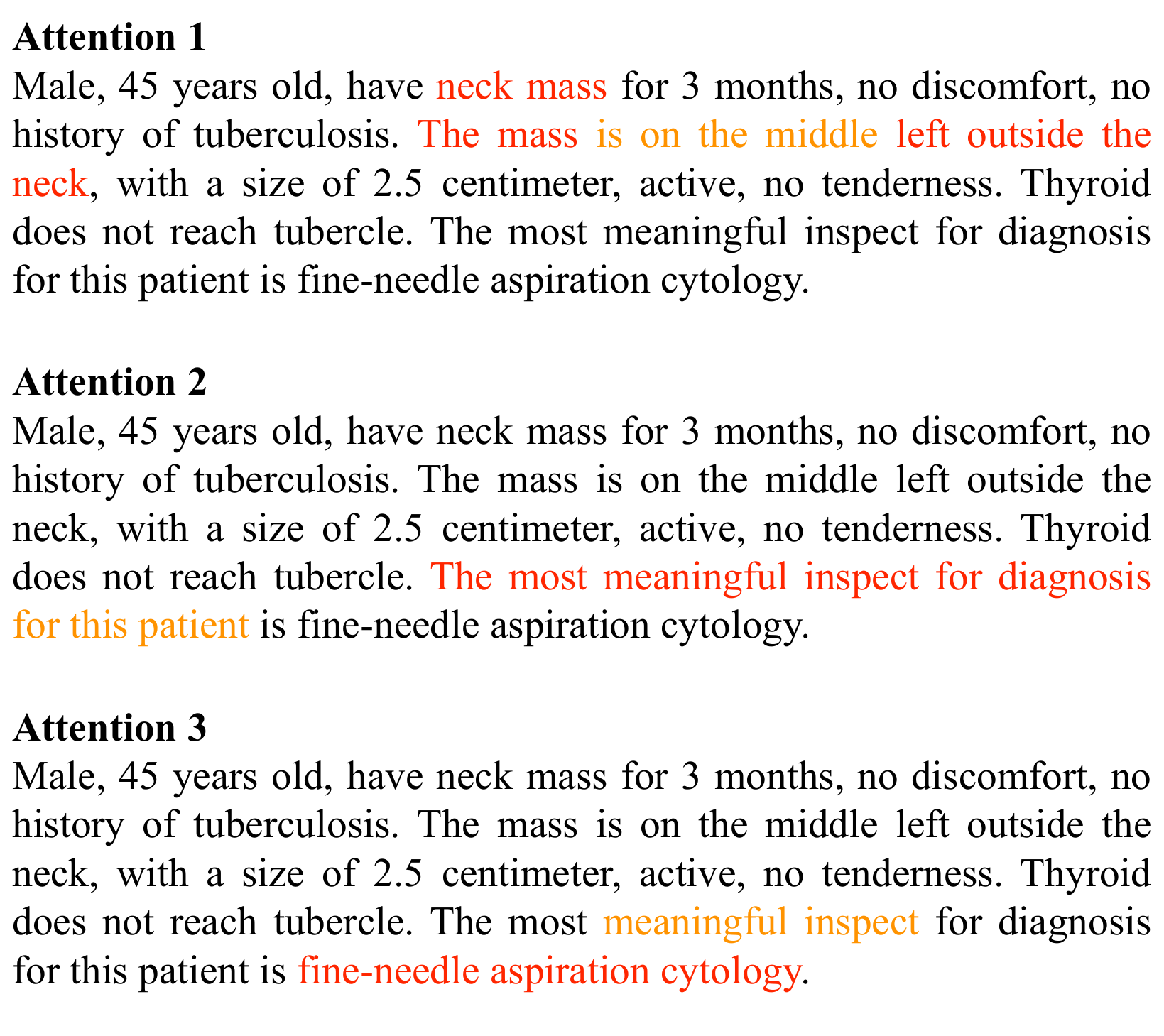}}
    \caption{An example of contextual self-attention over MedicalQA\#2 dataset. Red color indicates more attention value.}  
    \label{context_eg}    
\end{figure}
\begin{table}[h!]
\normalsize
    \centering
    \caption{Comparison with different self-attention strategies. MASE is the framework of Multi-scale self-Attention Sentence Embedding, which lacks context unit in the contextual self-attention layer. }
    \begin{tabular}{ccc}
        \hline
         Attention Strategy &MedicalQA\#1 & MedicalQA\#2 \\
         \hline 
         MASE &69.7 & 83.2\\ 
         CAMSE & \bf{73.6} & \bf{84.3} \\
         \hline 
    \end{tabular}
    \label{tab-atten-compare}
\end{table}


\begin{table}[h!]
\normalsize
    \centering
     \caption{Comparison with different scales in the framework CAMSE.}
    \begin{tabular}{ccc}
        \hline
         Multi-scale &MedicalQA\#1 & MedicalQA\#2 \\
         \hline 
         Conv 1  &72.1 & 83.9\\ 
         Conv 1+2 & 73.1 & 84.1 \\
         Conv 1+2+3 & \bf{73.6} & \bf{84.3} \\
         \hline 
    \end{tabular}
    \label{tab-conv-compare}
\end{table}

\subsection{Multi-scale layer}
\begin{figure}[htb!]        
    \center{\includegraphics[scale=0.5]  {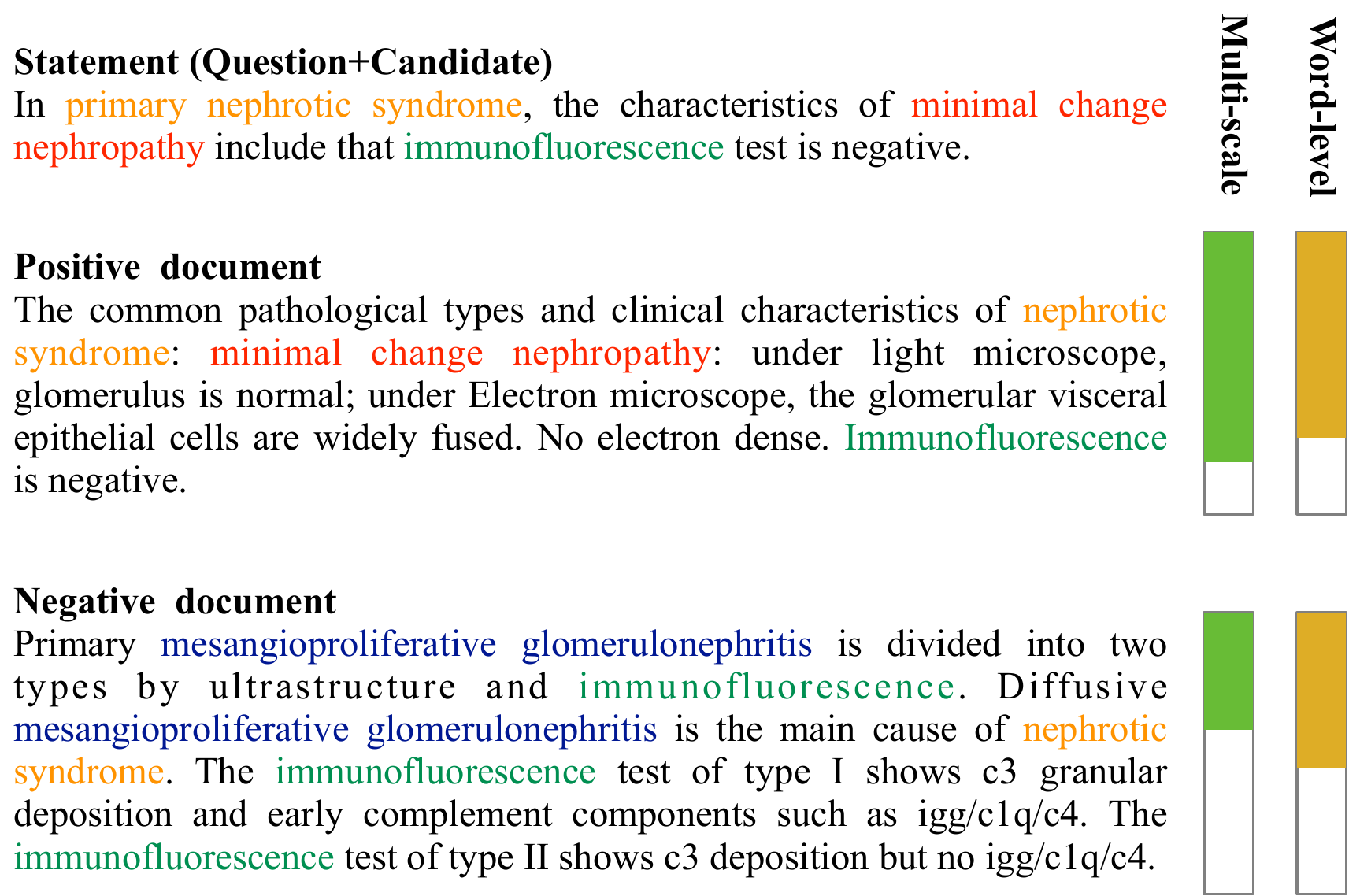}}
    \caption{An example of multi-scale layer over MedicalQA\#2 dataset.}  
    \label{Multi-scale}    
\end{figure}
The multi-scale layer aims at discovering entities, which is medical terminologies in our case. By binding sequential words, the multi-scale mechanism is capable of generating a representation for the entity and filter out noisy single words that also appear in the entity. 
Table \ref{tab-conv-compare} shows the improvement of multi-scale mechanism. We observe that the improvement is greater on MedicalQA\#1. The texts in MedicalQA\#1 are obtained from official examination and use lots of formal terminologies, which can be well captured by multi-scale layer, while the MedicalQA\#2 is subject to the diverse writing style of the doctors and harder to solve. 

Fig.\ref{Multi-scale} demonstrates how multi-scale mechanism outperforms the word-level models. The longer the strip is, the more support the statement receives from the document. Both the statement and positive document is about \emph{minimal change nephropathy}, a disease of kidney, while the negative document is about \emph{mesangioproliferative glomerulonephritis}, another disease of kidney. The colored keywords in statement, including \emph{primary nephrotic syndrome}, \emph{immunofluorescence}, are presented in both documents. And \emph{nephropathy} has a similar word embedding with \emph{glomerulonephritis}. Therefore, the word-level model would encounter difficulty when trying to distinguish these sentences with similar words. The noisy words in the negative document, such as \emph{immunofluorescence} and \emph{glomerulonephritis}, confuse the word-level model. However, the multi-scale model is able to recognize \emph{minimal change nephropathy} and \emph{mesangioproliferative glomerulonephritis} as integral units and easily discover that they are different diseases. The embeddings of them are distant enough so that the model can filter out irrelevant information from negative documents.

\subsubsection{SMS and SAS}

\begin{table}[h]
\normalsize
    \centering
    \caption{Comparison with using different scoring strategies.}
    \begin{tabular}{ccc}
        \hline
         Scoring Method &MedicalQA\#1 & MedicalQA\#2 \\
         \hline
        CAMSE(SAS + SMS) & \bf{73.6} & \bf{ 84.3}\\
         \hline
         CAMSE(SAS only) & 70.8 & 83.6 \\ 
         CAMSE(SMS only)  & 71.3 & 82.8 \\ 
        \hline
    \end{tabular}
    \label{tab-score-compare}
\end{table}

The SMS measures the similarity between sentences in aligned semantic sub-space, while the SAS strategy catches association across sub-space semantics. An example from CD-EMR illustrates how they function. The number of semantic sub-space is 5. We label the keywords of each sub-space, that is to say, words with highest attention intensity in each semantic. In parentheses are the number of sub-space the words contained in the brackets belong to. 

$\mathbf{Question:}$
[Rhinorrhea and expectoration](5) for 4 days. [Catch a cold](2) 4 days ago and then cough, produce expectoration, have [headache](4), [sore throat](1) and [rhinorrhea](3).

$\mathbf{Document:}$
[Nasal obstruction, cough and expectoration](5) for 3 days. [After catching a cold](2) 4 days ago, the patient began to cough, produce expectoration and [rhinorrhea](3), with suffering from headache and [sore throat](1). The patient has [dry stool](4).

Fig.\ref{score_eg} shows the  matrix of SMS(diagonal) and SAS(non-diagonal). The SMS approach directly compare similarities, thus semantic pairs $(Q1,D1)$, $(Q2,D2)$ and $(Q3,D3)$ have higher scores, while $(Q4,D4)$ and $(Q5,D5)$ are relatively lower. The SAS approach, in this case, manages to seize the association between (sore throat, rhinorrhea), (catch a cold, rhinorrhea), (catch a cold, headache), (rhinorrhea, nasal obstruction) respectively in semantic pairs$(Q1,D3)$, $(Q2,D3)$, $(Q4,D2)$ and $(Q3,D5)$. From a large scale of data, the model discovers symptom-symptom and cause-symptom association and using it to build up connections between different aspects of the description for a disease. 

\begin{figure}[ht!]        
    \center{\includegraphics[scale=0.2]  {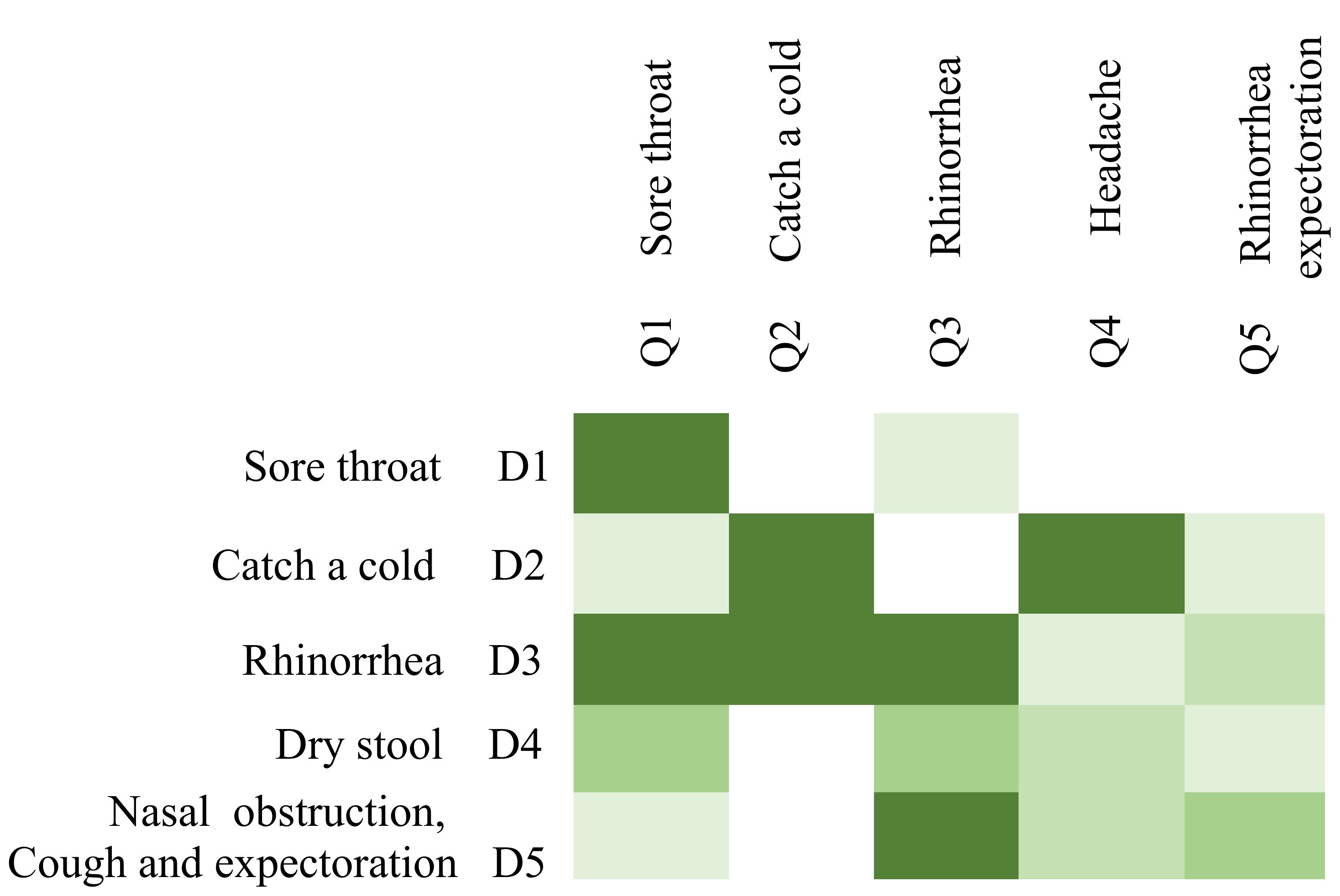}}
    \caption{An example of SMS and SAS.}  
    \label{score_eg}    
\end{figure}

\section{Conclusion}
In this paper, we introduce a kind of MedicalQA task and exploit sentence embedding for this problem.  A supervised learning module CAMSE is introduced to encode a sentence into an embedding tensor, and then two complementary scoring strategies Semantic Matching Scoring (SMS), and Semantic Association Scoring (SAS) are presented to exploit semantic similarity and association between a given question and the corresponding evidence document. 
   A series of experiments are conducted on two kinds of MedicalQA datasets to illustrate that our framework can achieve significantly better performance than competitive baseline approaches. Additionally, the proposed model can maintain better model interpretation with the contextual self-attention strategy to capture different semantic aspects at the sentence level. 

\section{ Acknowledgments}
We would like to thank Xiao Zhang for his help of  implementing some baseline models.  
This work is supported by the National Key Research and Development Program of China (No.2018YFC0116800).

\bibliographystyle{aaai}
\bibliography{CAMSE}
\end{document}